\documentclass[runningheads]{llncs}
\usepackage{graphicx}
\usepackage{url}
\usepackage{algorithm}
\usepackage{adjustbox}
\usepackage{amsmath}
\usepackage{amssymb}
\usepackage{listings}
\usepackage{adjustbox}
\usepackage{algpseudocode}
\usepackage{tabularx}
\usepackage{booktabs}
\usepackage{makecell}
\usepackage{placeins}
\usepackage{url}
\usepackage{tabularx}
\usepackage{cite}
\usepackage{enumitem}
\setlist{nosep}

\title{Optimizing Day-Ahead Energy Trading with Proximal Policy Optimization and Blockchain}
\titlerunning{Energy Trading with PPO and Blockchain}
\author{
  Navneet Verma\inst{1}\orcidID{0009-0005-1728-9562} \and
  Ying Xie\inst{1}\orcidID{0000-0002-8661-8877}
}

\institute{
  Kennesaw State University, USA \\
  \email{\{nverma@students,yxie2\}@kennesaw.edu}
}
\date{August 2025}
\authorrunning{NVerma, YXie}
\begin{document}

\maketitle
\begin{abstract}
The increasing penetration of renewable energy sources in day-ahead energy markets introduces challenges in balancing supply and demand, ensuring grid resilience, and maintaining trust in decentralized trading systems. This paper proposes a novel framework that integrates the Proximal Policy Optimization (PPO) algorithm, a state-of-the-art reinforcement learning method, with blockchain technology to optimize automated trading strategies for prosumers in day-ahead energy markets. We introduce a comprehensive framework that employs a Reinforcement Learning (RL) agent for multi-objective energy optimization and blockchain for tamper-proof data and transaction management. Simulations using real-world data from the Electricity Reliability Council of Texas (ERCOT) demonstrate the effectiveness of our approach. The RL agent achieves demand-supply balancing within 2\% of the demand and maintains near-optimal supply costs for the majority of the operating hours. Moreover, it generates robust battery storage policies capable of handling variability in solar and wind generation. All decisions are recorded on an Algorand-based blockchain, ensuring transparency, auditability, and security - key enablers for trustworthy multi-agent energy trading. Our key contributions are a novel system architecture, the use of curriculum learning to train the RL agent, and policy insights that support real-world deployment.
\end{abstract}

\section{Introduction}
The global energy landscape is undergoing a rapid transformation, driven by the increasing penetration of renewable energy sources, the decentralization of energy production, and the growing need for real-time, adaptive energy management strategies. Traditional centralized optimization approaches often fall short in managing the stochastic and dynamic behavior of modern power systems, particularly at the edge of the grid where uncertainty and variability are highest.

This paper explores the integration of Reinforcement Learning (RL) and Blockchain technology as a foundational approach for building intelligent, decentralized, and secure energy trading systems. RL provides a data-driven framework capable of learning optimal control policies through interaction with a dynamic environment, making it highly suitable for energy systems that involve variable renewables, shifting demand patterns, and uncertain market conditions.

\subsection{Motivation}
Traditionally Linear Programming (LP) and other mathematical optimization techniques have been widely used in energy management. These techniques typically rely on a centralized solver, require accurate modeling of all system parameters, and struggle with real-time adaptability. Furthermore, linear programming formulations must be resolved entirely whenever the problem parameters change beyond predefined limits. In contrast, RL is able to:
\begin{itemize}
\item Learn effectively from partial information and stochastic feedback.

\item Adapt to evolving system dynamics without needing to explicitly re-solve the problem once optimal policy is learned.  The majority of computational effort occurs only once during the training phase.

\item Handle multi-objective and sequential decision-making under uncertainty.

\end{itemize}

The deployment of RL agents at individual electric nodes — such as homes, substations, or microgrids — allows for local optimization that reflects site-specific conditions (e.g., solar irradiance, battery state-of-charge, local demand). This decentralized control paradigm offers several benefits.

\begin{itemize}

\item \textbf{Reduced Transmission Losses}: By optimizing generation and consumption locally, energy can be sourced and consumed closer to the point of use, minimizing line losses.

\item \textbf{Scalability}: Agents operate independently, avoiding the scalability limitations of centralized systems.

\textbf{Resilience}: Local autonomy enables the grid to continue functioning during partial failures or cyberattacks.

\end{itemize}

To ensure trust, transparency, and auditability in such a decentralized, multi-agent system, Blockchain technology plays a critical role. In our framework, an Algorand-based blockchain records agent actions, market transactions, and pricing decisions, providing:

\begin{itemize}
\item Immutable logs of transactions and agent decisions, enabling regulatory compliance and dispute resolution.

\item Tamper-proof coordination among agents without requiring centralized oversight.

\item Support for peer-to-peer trading through smart contracts and verifiable settlement mechanisms.
\end{itemize}

Together, RL and Blockchain technologies offer a powerful, synergistic foundation for the next generation of energy systems: autonomous, secure, and capable of continuous learning and adaptation.
\subsection{Contributions}
\noindent
\begin{minipage}{\linewidth}
This work offers several novel contributions.

\begin{enumerate}

\item \textbf{Novel Architecture}: Our work uses a hybrid RL-blockchain framework that integrates secure transaction management with multi-objective optimization using Proximal Policy Optimization (PPO) RL training algorithm. Unlike previous works relying on heuristic based optimization or MADDPG RL training algorithms, our approach leverages RL's capability in dynamic optimization and PPO’s stability and scalability for decentralized day ahead energy markets.
\item \textbf{Curriculum Based Learning}: Unlike traditional RL approaches that struggle with convergence in complex, high-variance environments, our framework employs curriculum-based learning to progressively train agents from simpler to more realistic scenarios. This strategy significantly improves training stability and policy robustness, particularly under the stochastic behavior of renewable energy sources and dynamic demand profile. To our knowledge, this is one of the first applications of curriculum learning in the context of decentralized energy trading, enabling faster learning and better generalization across diverse grid conditions.
\item	\textbf{Policy Recommendations}: Our work provides insights for regulators on integrating RL-blockchain systems into existing day ahead energy markets, addressing legal and technical barriers.
\item \textbf{Open-Source Implementation}: We provide a reference open source implementation of the framework, including RL algorithms and blockchain smart contracts, to facilitate further research and adoption.
\item \textbf{Interdisciplinary Approach}: We bridge machine learning, blockchain, and energy engineering to address multifaceted challenges in modern energy systems.

\end{enumerate}
\end{minipage}
\section{Literature Review}
The evolution of future energy trading systems will be guided by three fundamental principles: decarbonization, decentralization, and digitalization. However, the existing centralized structure of energy markets is ill-suited to achieve true decentralization. As highlighted by PwC \cite{PwC2025}, blockchain technology presents key advantages that can bridge this gap.

\begin{enumerate}
    \item \textbf{Decentralized Trading Platforms}: Blockchain enables peer-to-peer (P2P) energy trading by allowing energy producers—including residential and small-scale generators to transact directly without relying on a central intermediary. This approach enhances the market participation of individual stakeholders and promotes the development of innovative business models.
    \item \textbf{Grid Resilience}: The involvement of local participants enhances grid resilience by ensuring operational continuity and enabling rapid response to unexpected events. A simulation with an IEEE-123 node test feeder showed that P2P energy trading yielded 10.7\% improvement in resilience \cite{dwivedi2022}.
    \item \textbf{Power Loss Minimization}: Diverse studies, including (\cite{Azim2020}), have reported that P2P energy trading is one way to minimize long-distance transmission and distribution losses \cite{Vishwakarma2024}.
    \item \textbf{Transparency and Traceability}: Immutable records and transparent processes can significantly improve auditing and regulatory compliance \cite{DalCanto2017}.
\end{enumerate}

Al-Shehari et al. \cite{alshehari2024blockchain} integrated a heuristic optimization technique, the Mayfly Pelican Optimization Algorithm (MPOA), with blockchain to enable energy trading in the Internet of Electric Vehicles (IoEV). Similar to reinforcement learning approaches, MPOA seeks to balance exploration (global search) and exploitation (local refinement) when addressing non-convex optimization problems. The authors emphasize the importance of the computational efficiency and rapid convergence of the MPOA algorithm in addressing complex optimization problems, such as energy trading.

While heuristic methods such as MPOA \cite{alshehari2024blockchain} provide fast convergence for static optimization, they require repeated re-optimization in dynamic environments. In contrast, reinforcement learning can learn adaptive policies that account for temporal dependencies, uncertainty, and multi-agent interactions, making it more suitable for real-time energy trading.

Deep Reinforcement Learning (DRL) has proven to be a powerful approach for tackling complex decision-making challenges, particularly in the context of Electric Vehicle (EV) charging optimization within smart grids. By continuously interacting with the environment and updating its policies, DRL effectively adapts to uncertainties and fluctuating demand \cite{Han2024}. The study highlights that integrating a multi-agent deep reinforcement learning (MADRL) framework with blockchain significantly improves supply-demand balancing while ensuring secure transactions. However, the authors also emphasize the computational complexity involved in implementing the MADRL framework and managing cross-chain interactions when using the Hyperledger Fabric blockchain platform.

Xu et al. \cite{xu2020deep} integrated deep reinforcement learning with the Ethereum blockchain to enable secure peer-to-peer energy trading among microgrids. They modeled the utility maximization problem of each microgrid as a Markov game and employed a multi-agent deep deterministic policy gradient (MADDPG) algorithm to solve it. The authors noted that the presence of uncertainties and temporally coupled constraints associated with energy storage devices (batteries) makes it highly challenging to derive an optimal policy for each microgrid.

\section{Methodology}

In this work, we propose a methodology that combines Reinforcement Learning (RL) with blockchain technology to meet two essential requirements of modern energy systems: intelligent optimization and secure coordination. The RL component is employed to optimize energy flows by learning cost-effective and adaptive control policies, while the blockchain provides a decentralized and tamper-proof platform to ensure trust, transparency, and auditability of transactions. Together, these components form a resilient and scalable framework for efficient energy management.

\subsection{System Architecture}

\begin{figure}[ht]
\centering
\includegraphics[width=0.7\textwidth]{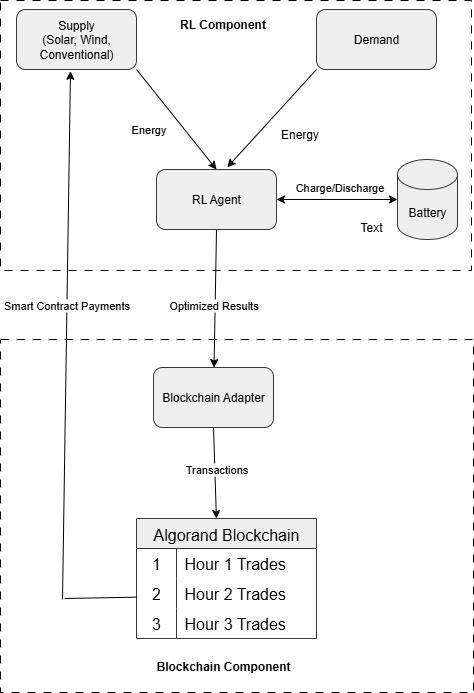}
\caption{High-level system architecture}
\label{fig:system_architecture}
\end{figure}

We adopt a layered architecture where the RL agent serves as the core domain logic and the blockchain is integrated via an intermediary adapter for the secure and auditable persistence of trading decisions. Figure~\ref{fig:system_architecture} presents the high-level system architecture.

\FloatBarrier

\subsubsection{Reinforcement Learning}
Many real-world tasks, such as making optimal trades or playing chess, lack labeled data and do not possess an explicit structure that can be directly exploited. Instead, learning occurs through interaction: the agent takes actions, observes the outcomes, and improves its behavior based on experience. Unlike supervised learning, where the correct answer is immediately available, the consequences of actions are often delayed. The primary objective is to learn policies that maximize cumulative rewards rather than simply fitting to data — a paradigm known as Reinforcement Learning (RL). The table below describes the terms and notations used in this paper.

\begin{table}[H]
\centering
\caption{RL terms and notations.}
\label{tab:rl-terms}
\begin{tabularx}{\columnwidth}{l c X}
\hline
\textbf{Term} & \textbf{Symbol} & \textbf{Meaning} \\ \hline
State & $s_i$ & Environment status at point i. \\
Action & $a_i$ & Decision at point i. \\
Reward & $r_i$ & Feedback for $a_i$. \\
Policy & $\pi_\theta(a|s)$ & Probabilistic decision rule. \\
Trajectory &$\tau$&Sequence of action and states. \\
Return & $R(\tau)$ & Expected return from trajectory $\tau$. \\
Episode Length & n & Length of full interaction sequence. \\
Discount & $\gamma$ & Weight for future rewards. \\
\hline
\end{tabularx}
\end{table}

\subsubsection{Markov Decision Process}
A Markov Decision Process (MDP) is a tuple (S, A, T, r), where
S is the set of states \[S = \{ s_i | i \in \mathbb{N}\} \]
A is the set of actions \[ A = \{ a_i | i \in \mathbb{N} \} \] 
T is the transition function\[ T: S\times A \times S -> [0, 1], \sum_{s_j}T(s_i, a, s_j) = 1\]
r is the reward function\[ r: S\times A  -> \mathbb{R}\]

We define trajectory $\tau$ as sequence of state, action and reward with n denoting the episode length.
\[\tau = \{ (s_0, a_0, r_0), (s_1, a_1, r_1), \dots, (s_{n-1}, a_{n-1}, r_{n-1}) \}\]

The return $R(\tau)$ for the trajectory $\tau$ is defined as the sum of discounted future rewards, $0 < \gamma < 1$.
\[R(\tau) = \sum_{i=0}^{n-1}{r_i*\gamma^i}\]

To get the optimal policy, we adjust the policy parameters such that expected value of the reward over the sampled trajectories is maximized.

\subsubsection{Proximal Policy Optimization}
To motivate the Proximal Policy Optimization (PPO) algorithm, we first summarize the vanilla policy gradient formulation on which PPO builds. The policy based reinforcement learning methods as described in \cite{Schulman2017Proximal} learn a policy $\pi_\theta(a|s)$ that selects actions 
a in each state s so as to maximize the expected total discounted reward. 

\begin{itemize}
    \item Sample multiple trajectories $\tau_1, \tau_2, \dots, \tau_k$ from the environment, each of fixed length $n$, where
    \[
        \tau_i = (s_0^{(i)}, a_0^{(i)}, s_1^{(i)}, a_1^{(i)}, \dots, s_{n}^{(i)}).
    \]
    
    \item The probability of observing a trajectory $\tau$ under policy $\pi_\theta$ is
    \begin{equation}
        p_\theta(\tau) = p(s_0) \prod_{t=0}^{n-1} \pi_\theta(a_t \mid s_t)\, p(s_{t+1} \mid s_t, a_t),
    \end{equation}
    where $p(s_0)$ is the probability of starting in the initial state $s_0$ and 
    $p(s_{t+1} \mid s_t, a_t)$ is the probability of transitioning from state $s_t$ to $s_{t+1}$ given action $a_t$.
    
    \item Express the objective function $J(\theta)$ as
    \begin{equation}
        J(\theta) = \frac{1}{k} \sum_{i=1}^{k} 
        \left( \sum_{t=0}^{n-1} 
        \log \pi_\theta\big(a_t^{(i)} \mid s_t^{(i)}\big) \right) 
        R(\tau_i),
    \end{equation}
    where $R(\tau_i)$ denotes the return of trajectory $\tau_i$.
    
    \item Apply a gradient ascent update to the parameters $\theta$ with learning rate $\alpha$:
    \begin{equation}
        \theta \leftarrow \theta + \alpha \, \nabla_\theta J(\theta).
    \end{equation}
\end{itemize}

Proximal Policy Optimization (PPO) improves upon the above formulation by making updates to the policy in small, controlled steps.

\subsection{Model Training and Optimization}
To accelerate convergence and improve policy stability, we adopt a curriculum learning strategy during training. Instead of exposing the agent to the full complexity of the environment from the start, the learning process begins with simpler tasks and gradually progresses to more challenging ones. This staged approach allows the policy to acquire basic decision-making skills early on and then refine them as task difficulty increases. In our setting, the curriculum is designed by incrementally adjusting environment parameters, such as target imbalance percentage and optimal cost thresholds, ensuring that the agent first masters easier scenarios before encountering more ambitious imbalance and optimal cost targets. Such progressive training has been shown to reduce variance in policy updates, leading to more robust and sample-efficient learning \cite{narvekar2020curriculum}. 

\subsection{Smart Contract Design}
The smart contract is designed to automate settlement of energy transactions between distributed energy resources (DERs) and consumers. By encoding market rules on-chain, it ensures transparent pricing, secure trade settlement, and tamper-proof record-keeping without requiring a central authority. The following table presents the key functional requirements for the smart contract.

\begin{table}[H]
\renewcommand{\arraystretch}{1.2}
\caption{Smart Contract Functional Requirements}
\label{tab:sc_requirements}
\centering
\begin{tabular}{p{4cm} p{6.2cm}}
\hline
\textbf{Requirement} & \textbf{Description} \\ \hline
Data Recording & Stores the trade settlement price and hourly timestamp in the Algorand global state to ensure immutable record-keeping. \\
Access Control & Only authenticated and registered participants on the Algorand blockchain can invoke contract functions. \\
Verification & Validates the submitted trade price and ensures no double-spending before execution. \\
Settlement & Automatically transfers Algorand tokens (Algos) to settle the trade after successful verification. \\
Error Handling & Fail gracefully when verification fails. \\ \hline
\end{tabular}
\end{table}

By integrating tamper proof data recording, robust verification mechanisms and graceful error handling, the proposed smart contract design fosters trust within a decentralized environment, thus facilitating the secure and reliable operation of autonomous agents engaged in energy trading on the grid.  Smart contract development on Algorand is simplified by PyTEAL, a Python-based high-level language that offers greater accessibility and faster prototyping compared to other blockchain platforms that rely on lower-level or less familiar languages. PyTEAL’s expressive syntax and integration with Python tools reduce the learning curve and accelerate secure smart contract implementation.

\subsection{Evaluation Metrics}
We evaluate the performance of RL agent by using the Imbalance Gap (\%) and Best Bound Gap (\%) metrics, defined below.
\begin{equation}
Imbalance\,Gap (\%) = \frac{\lvert Demand - Supply \rvert}{Demand} * 100
\end{equation}

\begin{equation}
Best\,Bound\,Gap (\%) = \frac{\lvert Actual\,Cost - Best Bound \rvert}{Actual\,Cost} * 100
\label{eq:CostGap}
\end{equation}

The best bound in equation \ref{eq:CostGap} represents an estimate of the minimum achievable cost, typically derived from the merit-order dispatch of generators. This estimate serves as a reference to guide the RL agent toward optimizing cost objectives. Alternative heuristics for estimating the best achievable cost, beyond merit-order dispatch, may also be employed.

For evaluating the performance of Algorand blockchain platform, we use the measures of transaction latency and throughput. Transaction latency is estimated by calculating the difference between the timestamps of the confirmed round and the first valid round of a transaction. In Algorand, a round refers to a discrete block produced by the network approximately every 4 seconds. The first valid round specifies the earliest block at which the transaction can be included, while the confirmed round indicates the block in which the transaction was actually finalized. By subtracting the timestamp of the first valid round from that of the confirmed round, we approximate the time taken for the transaction to be confirmed.Throughput measures the rate at which transactions are successfully processed and confirmed by the blockchain, typically expressed in transactions per second (txns/s). It is a key performance metric as it reflects the system’s capacity to handle high volumes of transactions, which is critical for scalability and for supporting real-time or large-scale decentralized applications.

\begin{equation}
\text{Latency (seconds)} = T_{\text{confirmation}} - T_{\text{first-valid}}
\label{eq:latency}
\end{equation}

where
\begin{align*}
T_{\text{confirmation}} & = \text{Timestamp of the round in which the transaction is confirmed}, \\
T_{\text{first-valid}} & = \text{Timestamp of the first round when the transaction becomes valid} \\
\end{align*}

\begin{equation}
\text{Throughput (txns/s)} = \frac{N_{\text{submitted}}}{L_{\text{avg}}}
\label{eq:throughput}
\end{equation}

where
\begin{align*}
N_{\text{submitted}} & = \text{Number of submitted transactions}, \\
L_{\text{avg}} & = \text{Average Latency} \\
\end{align*}
\section{Experimental Setup}
We provide a detailed description of the experimental framework, covering the dataset, hardware, and software configurations. The setup is designed to ensure reliable data processing, efficient model training, and well-structured RL reward formulation to achieve optimal performance. All end-to-end experiments involving interaction with the blockchain platform were conducted using a fixed random seed to ensure reproducibility. However, due to the inherent stochasticity of reinforcement learning algorithms, RL policy outcomes can vary across different seeds. To assess the robustness of the RL policy, we performed experiments with 10 distinct random seeds and averaged performance metrics to provide a reliable estimate of policy behavior while mitigating variability from individual runs. To incorporate input stochasticity across random seeds, we added Gaussian noise to system demand and prices, with mean 0 and standard deviations of 500 MW for demand and \$1/MWh for price.  

\subsection{Dataset}
Historical load and generation data from the Electric Reliability Council of Texas (ERCOT) were utilized to model realistic energy demand and supply profiles \cite{ercotdata2025}. These data serve as the basis for simulating market conditions and validating the proposed energy trading framework. To enhance the diversity of training scenarios, the original ERCOT data were slightly perturbed by introducing controlled random variations. This ensures that the RL agent is exposed to a range of operating conditions across different training episodes, improving its generalization capability.

\subsection{Software Configuration}
The system is implemented in Python 3.12.0, with comprehensive version control through our GitHub repository. The development environment supports both local and distributed computing configurations, optimized for machine learning tasks. Our implementation relies on the following specialized libraries and platforms:
\begin{itemize}
    \item \textbf{RL Algorithms}: Stable-Baselines3 2.6.0 for Reinforcement Learning algorithms such as PPO with pytorch 2.7.0+cpu
    \item \textbf{RL Environment Toolkit}: Gym 0.26.2 for modeling environment abstraction needed for reinforcement learning. Stable-Baselines3 interacts seamlessly with Gym environments by adhering to the standard Gym API (reset() and step() methods), allowing algorithms to remain agnostic to the specific environment dynamics. 
    \item \textbf{Blockchain Platform}: Algorand TestNet for on-chain data storage and smart contract deployment
    \item \textbf{Blockchain API}: The Algorand Python SDK was used for interfacing with Algorand platform. This lightweight API does not require running a local Algorand node, as transactions can be submitted via publicly available Algod or Indexer APIs, enabling rapid prototyping and deployment.
\end{itemize}

\subsection{Hardware Configuration}
\begin{table}[H]
\centering
\begin{tabular}{|c|c|c|}
\hline
\textbf{Component} & \textbf{Description} \\
\hline
OS Name & Microsoft Windows 11 Pro \\
CPU & 13th Gen Intel(R) Core(TM) i9-13900H, 14 cores\\
GPU 1 &	NVIDIA GeForce RTX 4060 \\
RAM &	64 GB  \\
GPU Memory &	8 GB  \\
\hline
\end{tabular}
\caption{Hardware Configuration}
\label{tab:hardware_config}
\end{table}
Automatic fallback to CPU is supported if GPU is not available.

\subsection{RL Agent Configuration}
\noindent
\begin{minipage}{\linewidth}
The RL agent is configured using the following parameters.
\begin{itemize}
    \item \textbf{State Space}: Solar and Wind profiles for current hour, current imbalance, best cost bound, demand and price forecast for current and next 6 hours.
    \item \textbf{Action Space}: Continuous actions representing fraction of available supply capacities for solar, wind, battery and conventional generators to match the energy demand in current hour.
    \item \textbf{Reward Design}: The reward function is designed to strongly penalize invalid actions, such as dispatching solar generation during nighttime or violating battery capacity limits (overflow or underflow). It simultaneously incentivizes battery charging and discharging when price forecasts are favorable. Convergence toward optimal imbalance and cost targets is facilitated by combining a carefully tuned episode termination (done) criterion with curriculum learning, enabling the agent to first master simpler targets before progressively tackling more challenging scenarios.
\end{itemize}

The following table shows the curriculum learning schedule for the RL agent. We first start with easier targets, which become increasingly difficult as training progresses.
\begin{table}[H]
\centering
\begin{tabular}{|c|c|c|}
\hline
\textbf{Imbalance Gap \%} & \textbf{Best Bound Gap \%} & \textbf{Timesteps} \\
\hline
40 & 40 & 40,000 \\
20 & 30 & 50,000 \\
10 & 20 & 60,000 \\
5 & 10 & 80,000 \\
2 & 10 & 100,000 \\
\hline
\end{tabular}
\caption{Curriculum Learning Schedule}
\label{tab:curriculum}
\end{table}

The repository containing the complete implementation and detailed documentation is available at \cite{EnergyTrader2025}.
\end{minipage}
\section{Results and Evaluation}
his section presents the experimental evaluation of the proposed RL-based energy trading framework. We begin by analyzing the patterns of input demand, price fluctuations, and renewable generation profiles to establish the context for system operation. Based on this understanding, we then evaluate the performance of the RL model in managing energy dispatch and trading decisions, and assess the scalability and effectiveness of the Algorand blockchain platform under varying operational conditions.

\begin{figure}[H]
\centering
\includegraphics[width=0.7\textwidth]{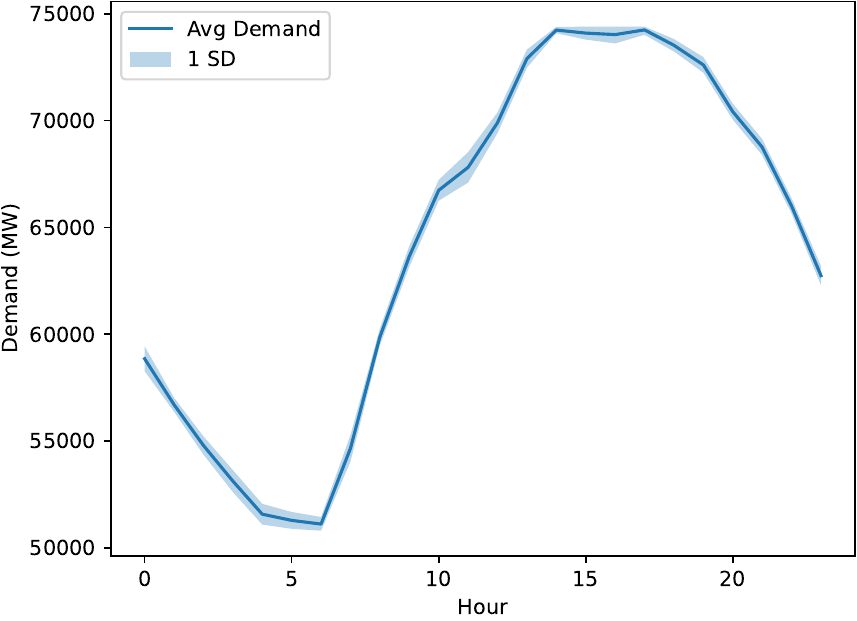}
\caption{System Demand}
\end{figure}
The demand follows a cyclical pattern, reaching its maximum in the evening hours. Prices closely mirror this behavior, with their peak occurring at the same time as the demand maximum. The shaded regions around the demand and price curves indicate the perturbations introduced by random noise across different seeds. 

\begin{figure}[H]
\centering
\includegraphics[width=0.7\textwidth]{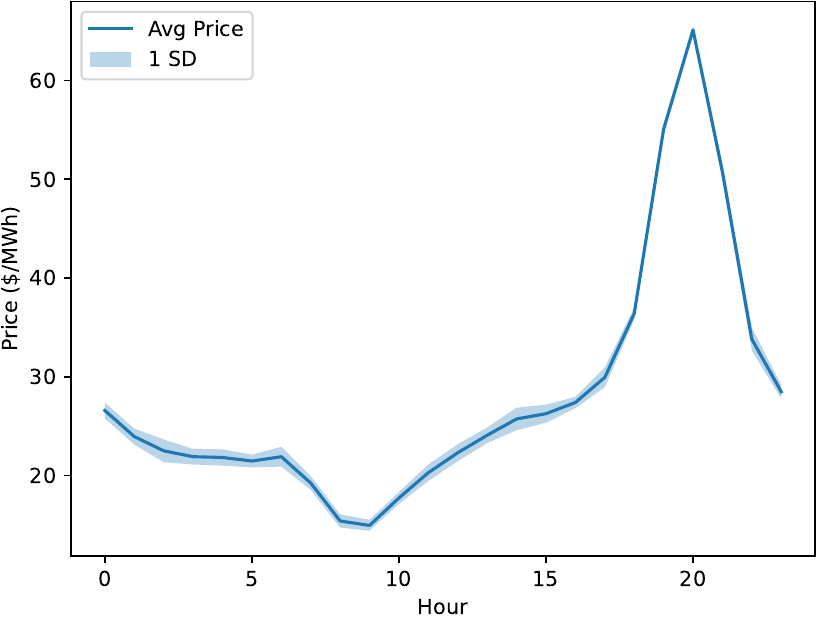}
\caption{Price}
\end{figure}
Prices range from \$14/MWh to \$65/MWh, with lows at night and highs during evening peaks. This spread creates arbitrage opportunities for battery storage, which the RL agent captures through its reward design.

\begin{figure}[H]
\centering
\includegraphics[width=0.7\textwidth]{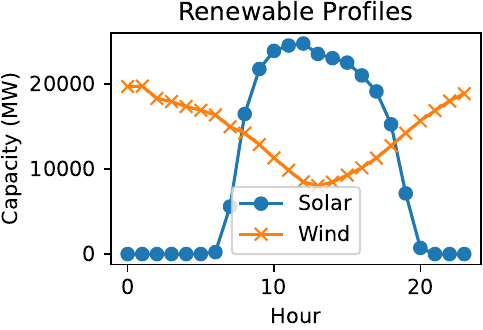}
\caption{Renewable Capacity Profiles}

\end{figure}
The renewable generation profile highlights its inherent variability, underscoring the critical role of battery operations in maintaining grid stability.

\subsection{RL Model Performance}
The following graphs depict the RL model’s performance metrics averaged over 10 distinct random seeds, with shaded regions representing one standard deviation. 

\begin{figure}[H]
\centering
\includegraphics[width=0.7\linewidth]{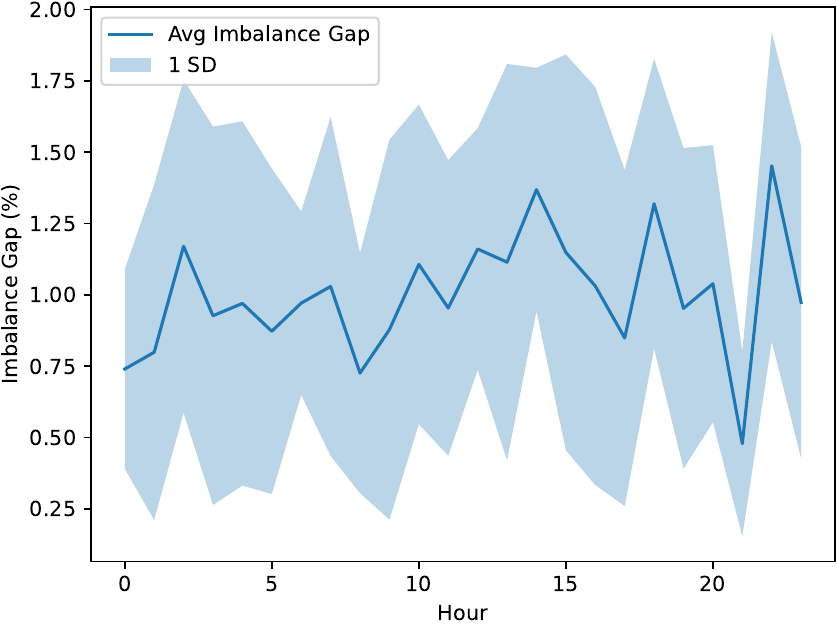}
\caption{Imbalance Gap (\%)}
\end{figure}
The RL model performs well to keep the supply-demand imbalance within 2\% of the demand.

\begin{figure}[H]
\centering
\includegraphics[width=0.7\textwidth]{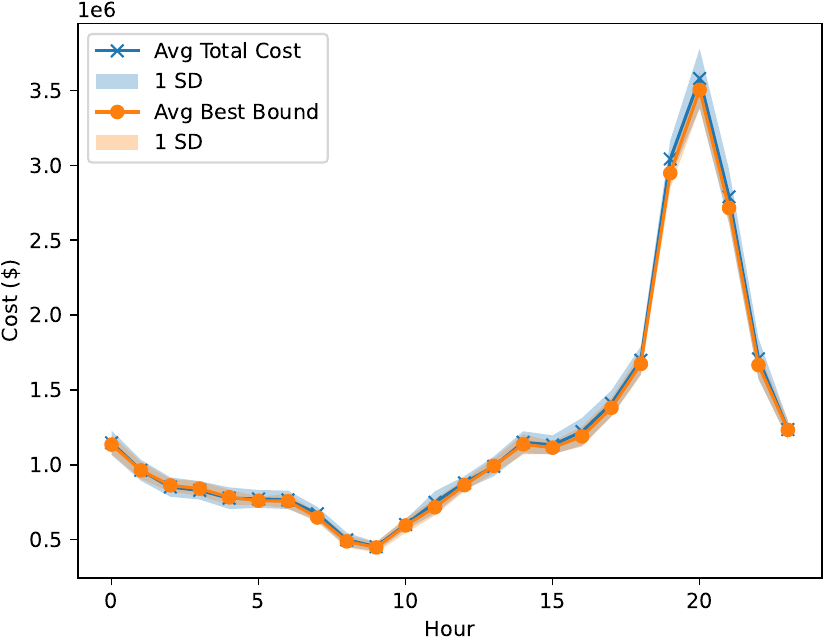}
\caption{Supply Cost vs Best Bound}
\end{figure}
The RL model achieves near-optimal costs across all hours, as supply costs closely track the best achievable values.

\begin{figure}[H]
\centering
\includegraphics[width=0.7\textwidth]{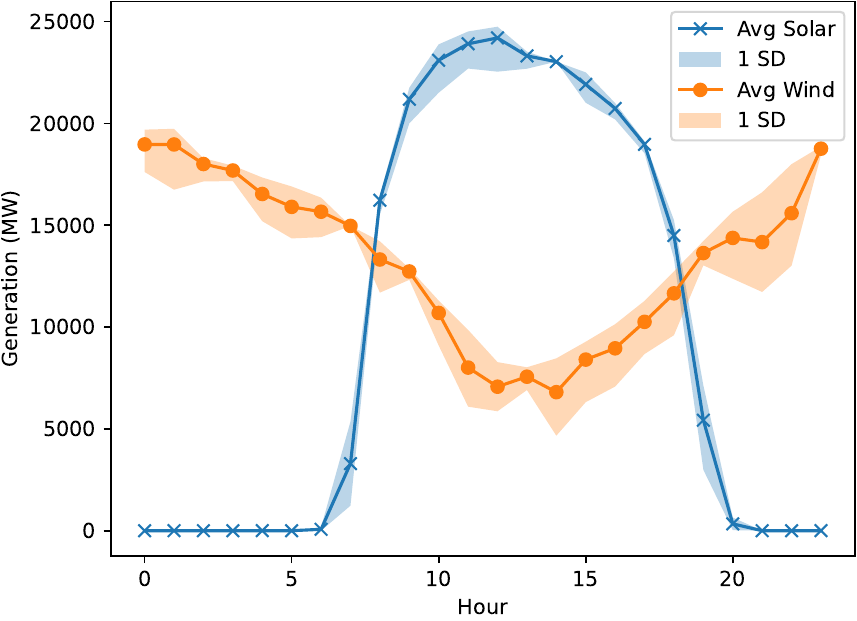}
\caption{Renewable Generation}
\end{figure}
The RL agent adheres to system constraints, utilizing renewable generation only when it is available.

\begin{figure}[H]
\centering
\includegraphics[width=0.7\textwidth]{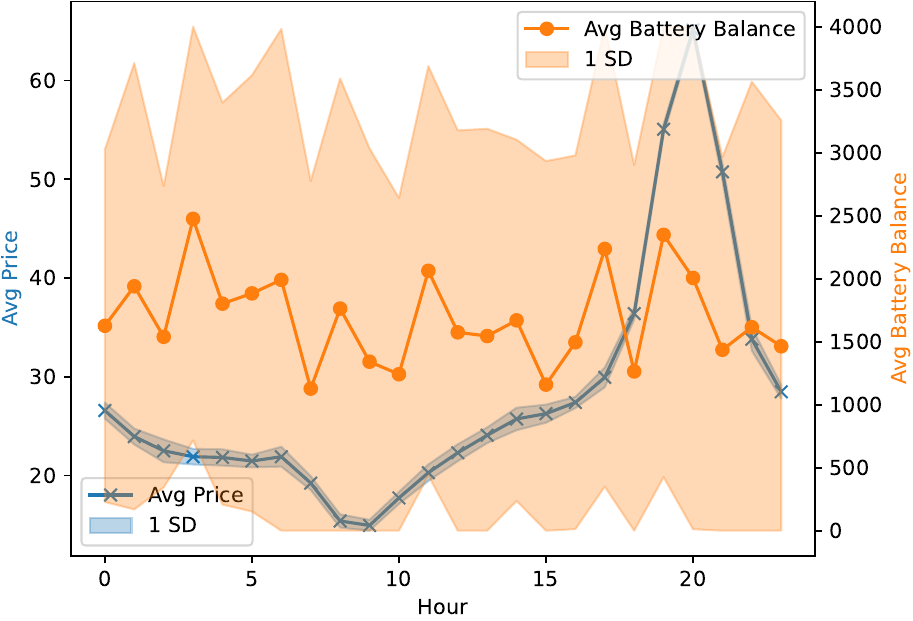}
\caption{Battery Balance}
\end{figure}
Frequent battery charge–discharge cycles result from the interaction of variable renewables, price arbitrage, and capacity limits. During rising peak prices, cycling maximizes short-term rewards, while in slightly declining periods, such as early mornings, it occurs to fully utilize renewable generation.

\subsection{Blockchain Performance}
On Algorand’s TestNet, we measured an average transaction confirmation latency of approximately 1.31 seconds across 203 transactions. Such performance is feasible in non-mainnet environments due to reduced validator participation and lower network congestion. The observed transaction throughput was approximately 155.51 transactions per second, constrained primarily by our blockchain adapter, which processed transactions sequentially. The Algorand network, however, is capable of supporting throughput up to 1000 transactions per second under parallel submission. Table \ref{tab:algorand_performance} summarizes these results.

\begin{table}[H]
\centering
\caption{Algorand TestNet Performance Metrics}
\label{tab:algorand_performance}
\begin{tabular}{|l|c|c|}
\hline
\textbf{Metric} & \textbf{Observed (TestNet)} & \textbf{Expected (MainNet)} \\ \hline
Average Latency & $\sim$1.31 s & 3-6 s \\ \hline
Throughput      & 155.51 txn/s & Up to 1000 txn/s \\ \hline
\end{tabular}
\end{table}
\FloatBarrier
\section{Conclusion and Future Work}
The study has explored integration of reinforcement learning with blockchain. The results show that reinforcement learning is effective in data driven multi-objective optimization for day ahead energy trading. Our work demonstrates that Algorand blockchain platform's low transaction latency, low transaction fees, and fast finality make it an ideal blockchain platform for energy trading. Its pure proof-of-stake consensus ensures scalability and security, enabling efficient, transparent, and trustless settlements in decentralized energy market. Although the results are promising, several aspects warrant further investigation. The first relates to regulatory compliance and practical deployment considerations, including market rules, participant consent, and verification of energy data. The second concerns the sensitivity of the agents’ performance to random seeds. These gaps are discussed in the following sub-sections, along with potential directions for improvement.

\subsection{Regulatory Compliance and Deployment Considerations}
Our system generates hourly energy trading transactions using reinforcement learning–based optimization, aiming to minimize imbalances between supply and demand at the lowest cost. Currently, participants are grouped into categories such as solar, wind, and traditional generators. Smart contracts manage category-level accounts, ensuring correct crediting and preventing double spending. However, preliminary user feedback highlights the importance of incorporating generator-specific characteristics—such as start-up and shut-down constraints, ramping limits, and offer prices that reflect actual production—for real-world deployment. In such a framework, payments would be directed to individual producers, with regulatory requirements, fees, and tariffs automatically applied.

We propose a hybrid design in which reinforcement learning allocates high-level categories, while unit commitment and economic dispatch models manage generator-level scheduling and settlement. This combines RL’s adaptability with the rigor of MILP, producing economically efficient, regulatorily compliant dispatch schedules suitable for blockchain-based settlement. At this stage, deployment on the Algorand testnet is sufficient for safe, low-cost validation, while mainnet deployment is envisioned for the future, representing a more realistic, regulator-approved implementation. As an additional implementation detail, blockchain-based verifiable reports could be included to further enhance transparency, trust, and dispute resolution for participants and regulators.

\subsection{Sensitivity to Random Seeds}
Although we conducted multiple experiments with different random seeds and found that on average, total system cost tracks well with the best estimate of system cost with average hourly gap varying between [0\% - 4.6\%], certain random seeds lead to higher deviations of up to 9.94\%. This variability is a well-known characteristic of policy gradient methods such as PPO, where random initialization, stochastic mini batch updates, and environment randomness can steer learning toward different local optima. In particular, sensitivity in the value function approximation and the variance of advantage estimates can amplify early differences between seeds, leading some runs to converge to less favorable solutions. Overall, the majority of seeds remain close to the mean performance, and the outliers reflect the inherent variance of PPO rather than a systematic issue in our setup.

\begin{figure}[H]
\centering
\includegraphics[width=0.7\textwidth]{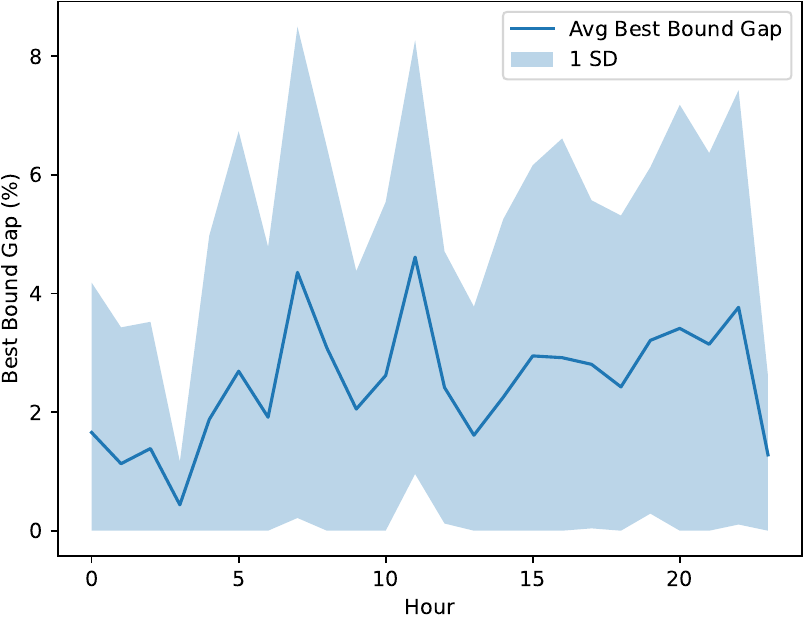}
\caption{Best Bound Gap}
\label{fig:BestBoundGap}
\end{figure}

As additional future work on the proposed hybrid model, we plan to integrate a model-based RL component to enhance policy robustness and short-horizon optimality. Unlike the current model-free RL, which updates policies immediately via trial-and-error, the model-based approach simulates multiple ‘what-if’ scenarios before updating. This involves learning a predictive environment model, simulating candidate actions over a planning horizon, and selecting actions that minimize cumulative predicted costs. Expected benefits include more stable learning and improved planning, reducing suboptimal short-term decisions.
\bibliographystyle{splncs04}
\clearpage
\bibliography{references}
\end{document}